\pdfoutput=1

\documentclass[11pt]{article}

\usepackage[]{naacl2021}

\usepackage{times}
\usepackage{latexsym}

\usepackage[T1]{fontenc}

\usepackage[utf8]{inputenc}

\usepackage{microtype}

\usepackage{booktabs}
\usepackage{colortbl}
\usepackage{caption}
\usepackage{subcaption}
\usepackage{graphicx}
\usepackage{float}
\usepackage{tikz}
\usetikzlibrary{shapes.misc, positioning}

\newcommand{\ASenti}{DynaSent}
\newcommand{\ModelZero}{Model~0}
\newcommand{\ModelOne}{Model~1}

\newcommand{\secref}[1]{Section~\ref{#1}}
\newcommand{\Figref}[1]{Figure~\ref{#1}}
\newcommand{\figref}[1]{Figure~\ref{#1}}
\newcommand{\Tabref}[1]{Table~\ref{#1}}
\newcommand{\tabref}[1]{Table~\ref{#1}}
\newcommand{\Appref}[1]{Appendix~\ref{#1}}
\newcommand{\appref}[1]{Appendix~\ref{#1}}

\definecolor{ourRed}{HTML}{E24A33}
\definecolor{ourBlue}{HTML}{348ABD}
\definecolor{ourPurple}{HTML}{988ED5}
\definecolor{ourGray}{HTML}{777777}
\definecolor{ourLightGray}{HTML}{B8B8B8}
\definecolor{ourYellow}{HTML}{FBC15E}
\definecolor{ourGreen}{HTML}{4D8951}
\definecolor{ourPink}{HTML}{FFB5B8}
\definecolor{oursteelblue}{HTML}{9BB8D7}
\definecolor{ourOrange}{HTML}{FDBA58}
\definecolor{ourWhite}{HTML}{FAFAFA}

\newcommand{\contribfootnote}[1]{%
  \begingroup
  \renewcommand{\thefootnote}{}\footnote{#1}%
  \addtocounter{footnote}{-1}%
  \endgroup
}

\newcommand{\totalExamples}{121,634}

\title{\ASenti: A Dynamic Benchmark for Sentiment Analysis}

\author{Christopher Potts$^{\ast}$ \\
  Stanford University \\
  \texttt{cgpotts@stanford.edu} \\\And
  Zhengxuan Wu$^{\ast}$  \\
  Stanford University \\
  \texttt{wuzhengx@stanford.edu}
  \\\AND
  Atticus Geiger \\
  Stanford University \\
  \texttt{atticusg@stanford.edu} \\\And
  Douwe Kiela \\
  Facebook AI Research \\
  \texttt{dkiela@fb.com}
}

\begin{document}
\maketitle

\begin{abstract}
  We introduce \ASenti\ (`Dynamic Sentiment'), a new English-language benchmark task for ternary~(positive/negative/neutral) sentiment analysis. \ASenti\ combines naturally occurring sentences with sentences created using the open-source Dynabench Platform, which facilities human-and-model-in-the-loop dataset creation.  \ASenti\ has a total of \totalExamples\ sentences, each validated by five crowdworkers, and its development and test splits are designed to produce chance performance for even the best models we have been able to develop; when future models solve this task, we will use them to create \ASenti\ version~2, continuing the dynamic evolution of this benchmark. Here, we report on the dataset creation effort, focusing on the steps we took to increase quality and reduce artifacts. We also present evidence that \ASenti's
  Neutral category is more coherent than the comparable category in other benchmarks, and we motivate training models from scratch for each round over successive fine-tuning.\contribfootnote{$^{\ast}$Equal contribution.}
\end{abstract}

\section{Introduction}\label{sec:intro}

Sentiment analysis is an early success story for NLP, in both a technical and an industrial sense. It has, however, entered into a more challenging phase for research and technology development: while present-day models achieve outstanding results on all available benchmark tasks, they still fall short when deployed as part of real-world systems~\citep{Burn-Murdoch:2013,Grimes:2014,Grimes:2017,Gossett:2020} and display a range of clear shortcomings \citep{kiritchenko-mohammad-2018-examining,HanwenShen-etal:2019,wallace-etal-2019-universal,tsai-etal-2019-adversarial-attack,jin2019bert,Zhang-etal:2020}.

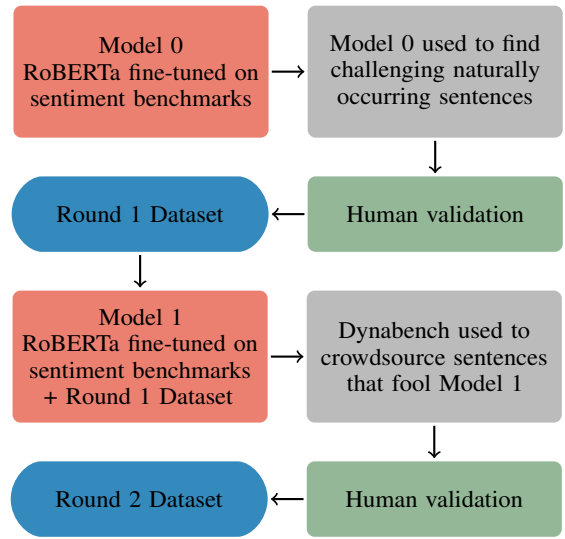
\begin{figure}[ht]
  \centering
  \newcommand{\repgap}{0.5cm}
\newcommand{\dynaText}[2]{#1\\#2}

\begin{tikzpicture}
  \tikzset{
    every node=[
    draw,
    font=\footnotesize,
    shape=rectangle,
    rounded corners=3pt,
    align=center,
    minimum width=3.1cm,
    minimum height=1cm,
    text width=3.1cm
    ],
    model0/.style={
      fill=ourRed!70,
      minimum height=1.75cm
    },
    harvest/.style={
      fill=ourGray!50,
      minimum height=1.75cm
    },
    validation/.style={
      fill=ourGreen!60
    },
    round1/.style={
      fill=ourBlue,
      shape=rounded rectangle
    },
    model1/.style={
      fill=ourRed!70,
      minimum height=1.75cm
    },
    dynabench/.style={
      fill=ourGray!50,
      minimum height=1.75cm
    },
    round2/.style={
      fill=ourBlue,
      shape=rounded rectangle
    },
    dynasentArrow/.style={
      solid,
      thick,
      ->,
      shorten <=1pt,
      shorten >=1pt
    }
  };

  \node[model0](model0){\dynaText{\ModelZero}{RoBERTa fine-tuned on sentiment benchmarks}};
  \node[harvest, right=\repgap of model0](harvest){\ModelZero\ used to find challenging naturally occurring sentences};
  \node[validation, below=\repgap of harvest](val1){Human validation};
  \node[round1, left=\repgap of val1](round1){Round 1 Dataset};
  \node[model1, below=\repgap of round1](model1){\dynaText{\ModelOne}{RoBERTa fine-tuned on sentiment benchmarks + Round 1 Dataset}};
  \node[dynabench, right=\repgap of model1](dynabench){Dynabench used to crowdsource sentences that fool \ModelOne};
  \node[validation, below=\repgap of dynabench](val2){Human validation};
  \node[round2, left=\repgap of val2](round2){Round 2 Dataset};

  \path(model0.east) edge[dynasentArrow] (harvest.west);
  \path(harvest.south) edge[dynasentArrow] (val1.north);
  \path(val1.west) edge[dynasentArrow] (round1.east);
  \path(round1.south) edge[dynasentArrow] (model1.north);
  \path(model1.east) edge[dynasentArrow] (dynabench.west);
  \path(dynabench.south) edge[dynasentArrow] (val2.north);
  \path(val2.west) edge[dynasentArrow] (round2.east);

\end{tikzpicture}
  \caption{The \ASenti\ dataset creation process. The human validation task is the same for both rounds; five responses are obtained for each sentence. On Dynabench, we explore conditions with and without prompt sentences that workers can edit to achieve their goal.}
  \label{fig:overview}
\end{figure}

In this paper, we seek to address the gap between benchmark results and actual utility by introducing version~1 of the \ASenti\ dataset for English-language ternary (positive/negative/neutral) sentiment analysis.\footnote{\url{https://github.com/cgpotts/dynasent}} \ASenti\ is intended to be a \emph{dynamic} benchmark that expands in response to new models, new modeling goals, and new adversarial attacks. We present the first two rounds here and motivate some specific data collection and modeling choices, and we propose that, when future models solve these rounds, we use those models to create additional \ASenti\ rounds. This is an instance of ``the `moving post' dynamic target'' for NLP that \citet{nie-etal-2020-adversarial} envision.

\Figref{fig:overview} summarizes our method, which incorporates both naturally occurring sentences and sentences created by crowdworkers with the goal of fooling a top-performing sentiment model. The starting point is \ModelZero, which is trained on standard sentiment benchmarks and used to find challenging sentences in existing data. These sentences are fed into a human validation task, leading to the Round~1 Dataset. Next, we train \ModelOne\ on Round~1 data in addition to publicly available datasets. In Round 2, this model runs live on the Dynabench Platform for human-and-model-in-the-loop dataset creation;\footnote{\url{https://dynabench.org/}} crowdworkers try to construct examples that fool \ModelOne. These examples are human-validated, which results in the Round~2 Dataset. Taken together, Rounds 1 and 2 have \totalExamples\ sentences, each with five human validation labels. Thus,  with only two rounds collected, \ASenti\ is already a substantial new resource for sentiment analysis.

In addition to contributing \ASenti, we seek to address a pressing concern for any dataset collection method in which workers are asked to construct original sentences: human creativity has intrinsic limits. Individual workers will happen upon specific strategies and repeat them, and this will lead to dataset artifacts. These artifacts will certainly reduce the value of the dataset, and they are likely to perpetuate and amplify social biases.

We explore two methods for mitigating these dangers. First, by harvesting naturally occurring examples for Round~1, we tap into a wider population than we can via crowdsourcing, and we bring in sentences that were created for naturalistic reasons \citep{devries2020towardsecologically}, rather than the more artificial goals present during crowdsourcing. Second, for the Dynabench cases created in Round~2, we employ a `Prompt' setting, in which crowdworkers are asked to modify a naturally occurring example rather than writing one from scratch. We compare these sentences with those created without a prompt, and we find that the prompt-derived sentences are more like naturally occurring sentences in length and lexical diversity. Of course, fundamental sources of bias remain -- we seek to identify these in the Datasheet \citep{gebru2018datasheets} distributed with our dataset -- but we argue that these steps help, and can inform crowdsourcing efforts in general.

As noted above, \ASenti\ presently uses the labels Positive, Negative, and Neutral. This is a minimal expansion of the usual binary (Positive/Negative) sentiment task, but a crucial one, as it avoids the false presupposition that all texts convey binary sentiment. We chose this version of the problem to show that even basic sentiment analysis poses substantial challenges for our field. We find that the Neutral category is especially difficult. While it is common to synthesize such a category from middle-scale product and service reviews, we use an independent validation of the Stanford Sentiment Treebank \citep{socher-etal-2013-recursive} dev set to argue that this tends to blur neutrality together with mixed sentiment and uncertain sentiment (\secref{sec:neutral}). \ASenti\ can help tease these phenomena apart, since it already has a large number of Neutral examples and a large number of examples displaying substantial variation in validation. Finally, we argue that the variable nature of the Neutral category is an obstacle to fine-tuning (\secref{sec:fine-tuning}), which favors our strategy of training models from scratch for each round.

\section{Related Work}

Sentiment analysis was one of the first natural language understanding tasks to be revolutionized by data-driven methods. Rather than trying to survey the field (see \citealt{PangLee08,Liu:2012,Grimes:2014}), we focus on the benchmark tasks that have emerged in this space, and then seek to situate these benchmarks with respect to challenge (adversarial) datasets and crowdsourcing methods.

\subsection{Sentiment Benchmarks}

The majority of sentiment datasets are derived from customer reviews of products and services. This is an appealing source of data, since such texts are accessible and abundant in many languages and regions of the world, and they tend to come with their own author-provided labels (star ratings). On the other hand, over-reliance on such texts is likely also limiting progress; \ASenti\ begins moving away from such texts, though it remains rooted in this domain.

\citet{pang-lee-2004-sentimental} released a collection of 2,000 movie reviews with binary sentiment labels. This dataset became one of the first benchmark tasks in the field, but it is less used today due to its small size and issues with its train/test split that make it artificially easy \citep{maas-etal-2011-learning}.

A year later, \citet{pang-lee-2005-seeing} released a  collection of 11,855 sentences extracted from movie reviews from the website Rotten Tomatoes. These sentences form the basis for the Stanford Sentiment Treebank (SST; \citealt{socher-etal-2013-recursive}), which provides labels (on a five-point scale) for all the phrases in the parse trees of the sentences in this dataset.

\citet{maas-etal-2011-learning} released a sample of 25K labeled movie reviews from IMDB, drawn from 1--4-star reviews (negative) and 7--10-star reviews (positive). This remains one of largest benchmark tasks for sentiment; whereas much larger datasets have been introduced into the literature at various times \citep{Jindal:Liu:2008,ni-etal-2019-justifying}, many have since been removed from public distribution~\citep{McAuley:Leskovec:Jurafsky:2012,Zhang-etal:2015}. A welcome exception is the Yelp Academic Dataset,\footnote{\url{https://www.yelp.com/dataset}} which was first released in 2010 and which has been expanded a number of times, so that it how has over 8M review texts with extensive metadata.

Not all sentiment benchmarks are based in review texts. The MPQA Opinion Corpus of \citet{Wiebe:Wilson:Cardie:2005} contains news articles labeled at the phrase-level by experts for a wide variety of subjective states; it presents an exciting vision for how sentiment analysis might become more multidimensional. SemEval 2016 and 2017 \citep{nakov-etal-2016-semeval,rosenthal-etal-2017-semeval} offered Twitter-based sentiment datasets that continue to be used widely despite their comparatively small size ($\approx$10K examples). And of course there are numerous additional datasets for specific languages, domains, and emotional dimensions; Google's Dataset Search currently reports over 100 datasets for sentiment, and the website Papers with Code gives statistics for dozens of sentiment tasks.\footnote{\url{https://paperswithcode.com/task/sentiment-analysis}}

Finally, sentiment lexicons have long played a central role in the field. They are used as resources for data-driven models, and they often form the basis for simple hand-crafted decision rules. The Harvard General Inquirer \citep{stone1963computer} is an early and still widely used multidimensional sentiment lexicon. Other important examples are OpinionFinder \citep{wilson-etal-2005-recognizing}, LIWC \citep{LIWC:2007}, MPQA, Opinion Lexicon \citep{OpinionLexicon}, SentiWordNet \citep{BaccianellaEsuliSebastiani10}, and the valence and arousal ratings of \citet{Warriner:2013aa}.

\subsection{Challenge and Adversarial Datasets}

Challenge and adversarial datasets have risen to prominence over the last few years in response to the sense that benchmark results are over-stating the quality of the models we are developing~\citep{linzen-2020-accelerate}. The general idea traces to \citet{Winograd:1972}, who proposed minimally contrasting examples meant to evaluate the ability of a model to capture specific cognitive, linguistic, and social phenomena (see also \citealt{Levesque:2013}). This guiding idea is present in many recent papers that seek to determine whether models have met specific learning targets  \citep{alzantot-etal-2018-generating,glockner-etal-2018-breaking,naik-etal-2018-stress,nie2019analyzing}. Related efforts reveal that models are exploiting relatively superficial properties of the data that are easily exploited by adversaries \citep{jia-liang-2017-adversarial,kaushik-lipton-2018-much,Zhang-etal:2020}, as well as social biases in the data they were trained on \citep{kiritchenko-mohammad-2018-examining,rudinger-etal-2017-social,rudinger-etal-2018-gender,sap-etal-2019-risk,schuster-etal-2019-towards}.

For the most part, challenge and adversarial datasets are meant to be used primarily for evaluation (though \citet{liu-etal-2019-inoculation} show that even small amounts of training on them can be fruitful in some scenarios).  However, there are existing adversarial datasets that are large enough to support full-scale training efforts \citep{zellers-etal-2018-swag,zellers-etal-2019-hellaswag,chen-etal-2019-codah,dua-etal-2019-drop,bartolo-etal-2020-beat}. \ASenti\ falls into this class; it has large train sets that can support from-scratch training as well as fine-tuning. Our approach is closest to, and directly inspired by, the Adversarial NLI (ANLI) project, which is reported on by \citet{nie-etal-2020-adversarial} and which continues on Dynabench. In ANLI, human annotators construct new examples that fool a top-performing model but make sense to other human annotators. This is an iterative process that allows the annotation project itself to organically find phenomena that fool current models. The resulting dataset has, by far, the largest gap between estimated human performance and model accuracy of any benchmark in the field right now. We hope \ASenti\ follows a similar pattern, and that its naturally occurring sentences and prompt-derived sentences bring beneficial diversity.

\begin{table*}[tp]
  \centering
  \newcommand{\spacer}{\hspace{28pt}}
  \begin{tabular}[c]{
        l 
        *{2}{r} @{\spacer} 
        *{2}{r} @{\spacer} 
        *{2}{r} 
    }
    \toprule
    & \multicolumn{2}{c@{\spacer}}{SST-3} 
    & \multicolumn{2}{c@{\spacer}}{Yelp} 
    & \multicolumn{2}{c}{Amazon} \\
    & Dev & Test & Dev & Test  & Dev & Test \\
    \midrule    
    Positive  &   444 &   909   &    9,577 & 10,423   & 130,631 & 129,369 \\
    Negative  &   428 &   912   &   10,222 &  9,778   & 129,108 & 130,892 \\
    Neutral   &   228 &   389   &    5,201 &  4,799   &  65,261 &  64,739  \\[1ex]
    Total     & 1,100 & 2,210   &   25,000 & 25,000   & 325,000 & 325,000 \\
    \bottomrule    
  \end{tabular}
  \caption{External assessment datasets used for \ModelZero\ and \ModelOne. SST-3 is the ternary version of SST as described in \tabref{tab:model0-train-data} but with only the sentence-level phrases included. For the Yelp and Amazon datasets, we split the test datasets of \citealt{Zhang-etal:2015} in half by line number to create a dev/test split.}
  \label{tab:external-assess-data}
\end{table*}

\subsection{Crowdsourcing Methods}\label{sec:crowdsourcing}

Within NLP, \citet{snow-etal-2008-cheap} helped establish crowdsourcing as a viable method for collecting data for at least some core language tasks. Since then, it has become the dominant mode for dataset creation throughout all of AI, and the scientific study of these methods has in turn grown rapidly. For our purposes, a few core findings from research into crowdsourcing are centrally important.

First, crowdworkers are not fully representative of the general population: they are, for example, more educated, more male, more technologically enabled \citep{Hube-etal:2019}. As a result, datasets created using these methods will tend to inherit these biases, which will then find their way into our models. \ASenti's naturally occurring sentences and prompt sentences can help, but we acknowledge that those texts come from people who write online reviews, which is also a special group.

Second, any crowdsourcing project will reach only a small population of workers, and this will bring its own biases. The workers will tend to adopt similar cognitive patterns and strategies that might not be representative even of the specialized population they belong to \citep{Gadiraj-etal:2017}. This seems to be an underlying cause of many of the artifacts that have been identified in prominent NLU benchmarks \citep{poliak-etal-2018-hypothesis,gururangan-etal-2018-annotation,tsuchiya-2018-performance,belinkov-etal-2019-dont}.

Third, as with all work, quality varies across workers and examples, which raises the question of how best to infer individual labels from response distributions. \citet{Dawid:Skene:1979} is an early contribution to this problem leveraging Expectation Maximization \citep{dempster1977maximum}. Much subsequent work has pursued similar strategies; for a full review, see \citealt{Zheng-etal:2017}. Our corpus release uses the true majority (3/5 labels) as the gold label where such a majority exists, leaving examples unlabeled otherwise, but we include the full response distributions in our corpus release and make use of those distributions when training \ModelOne. For additional details, see \secref{sec:round1-validation}.

\section{Round 1: Naturally Occurring Sentences}\label{sec:round1}

We now begin to describe our method for constructing \ASenti\ (\figref{fig:overview}). We begin with an initial model, \ModelZero, and use it to harvest challenging examples from an existing corpus. These examples are human-validated and incorporated into the training of a second model, \ModelOne. This model is then loaded into Dynabench; Round~2 of our data collection (\secref{sec:round2}) involves crowdworkers seeking to create examples that fool this model.

\begin{table*}[tp]
  \centering  
  \begin{subtable}[b]{1\textwidth}
    \setlength{\tabcolsep}{14pt}
    \centering
    \begin{tabular}[b]{l *{5}{r}}
      \toprule
      & CR    & IMDB   & SST-3 & Yelp & Amazon \\
      \midrule
      Positive & 2,405 & 12,500 & 42,672 & 260,000 & 1,200,000 \\
      Negative & 1,366 & 12,500 & 34,944 & 260,000 & 1,200,000 \\
      Neutral  & 0     & 0      & 81,658 & 130,000 & 600,000 \\[1ex]
      Total    & 3,771 & 25,000 & 159,274 & 650,000 & 3,000,000 \\
      \bottomrule
    \end{tabular}
    \caption{\ModelZero\ training data. CR is the Customer Reviews dataset from \citealt{HuLiu04}, IMDB is from \citealt{maas-etal-2011-learning}, SST-3 is the phrase-level ternary SST (labels 0--1 = Neg; 2 = Neu; 3--4 = Pos), and Yelp and Amazon are from \citealt{Zhang-etal:2015} (1--2-star reviews = Neg; 3-star = Neutral; 4--5-star = Pos).}
    \label{tab:model0-train-data}
  \end{subtable}
  
  \vspace{6pt}
  
  \begin{subtable}[b]{1\textwidth}
    \centering
    \newcommand{\spacer}{\hspace{28pt}}
    \begin{tabular}[b]{l *{2}{r} @{\spacer} *{2}{r} @{\spacer} *{2}{r} @{\spacer} *{2}{r} @{\spacer} *{2}{r} }
      \toprule
      & \multicolumn{2}{c@{\spacer}}{SST-3} 
      & \multicolumn{2}{c@{\spacer}}{Yelp} 
      & \multicolumn{2}{c@{\spacer}}{Amazon} 
      & \multicolumn{2}{c@{\spacer}}{Round~1} 
      & \multicolumn{2}{c}{Round~2}\\
      & Dev & Test & Dev & Test  & Dev & Test & Dev & Test & Dev & Test \\
      \midrule
      Positive  & 85.1 & 89.0 & 88.3 & 90.5 & 89.1 & 89.4 & 33.3 & 33.3 & 58.4 & 63.0 \\
      Negative  & 84.1 & 84.1 & 88.8 & 89.1 & 86.6 & 86.6 & 33.3 & 33.3 & 61.0 & 63.1 \\
      Neutral   & 45.4 & 43.5 & 58.2 & 59.4 & 53.9 & 53.7 & 33.3 & 33.3 & 38.4 & 44.3 \\[1ex]
      Macro avg & 71.5 & 72.2 & 78.4 & 79.7 & 76.5 & 76.6 & 33.3 & 33.3 & 52.6 & 56.8 \\
      \bottomrule
    \end{tabular}
    \caption{\ModelZero\ performance (F1 scores) on external assessment
      datasets (\tabref{tab:external-assess-data}). We also report on our Round~1 dataset (\secref{sec:round1-dataset}), where performance is at chance by construction, and we report on our Round~2 dataset (\secref{sec:round2}) to further quantify the challenging nature of that dataset.}
    \label{tab:model0-assess}
  \end{subtable}
  \caption{\ModelZero\ summary.}
  \label{tab:model0}
\end{table*}

\subsection{\ModelZero}\label{sec:model0}

Our \ModelZero\ begins with the RoBERTa-base parameters \citep{liu2019roberta} and adds a three-way sentiment classifier head. The model was trained on a number of publicly-available datasets, as summarized in \tabref{tab:model0-train-data}. The Customer Reviews \citep{HuLiu04} and IMDB \citep{maas-etal-2011-learning} datasets have only binary labels. The other datasets have five star-rating categories. We bin these ratings by taking the lowest two ratings to be negative, the middle rating to be neutral, and the highest two ratings to be positive. The Yelp and Amazon datasets are those used in \citealt{Zhang-etal:2015}; the first is derived from an earlier version of the Yelp Academic Dataset, and the second is derived from the dataset used by \citet{McAuley:Leskovec:Jurafsky:2012}. SST-3 is the ternary version of the SST; we train on the phrase-level version of the dataset (and always evaluate only on its sentence-level labels). For additional details on how this model was optimized, see \appref{app:model0}.

We evaluate this and subsequent models on three datasets (\tabref{tab:external-assess-data}): SST-3 dev and test, and the assessment portion of the Yelp and Amazon datasets from \citealt{Zhang-etal:2015}. For Yelp and Amazon, the original distribution contained only (very large) test files. We split them in half (by line number) to create dev and test splits.

In \Tabref{tab:model0-assess}, we summarize our \ModelZero\ assessments on these datasets. Across the board, our model does extremely well on the Positive and Negative categories, and less well on Neutral. We trace this to the fact that the Neutral categories for all these corpora were derived from three-star reviews, which actually mix a lot of different phenomena: neutrality, mixed sentiment, and (in the case of the reader judgments in SST) uncertainty about the author's intentions. We return to this issue in \secref{sec:neutral}, arguing that \ASenti\ marks progress on creating a more coherent Neutral category.

Finally, \tabref{tab:model0-assess} includes results for our Round~1 dataset, as we are defining it. Performance is at-chance across the board by construction (see \secref{sec:round1-dataset} below). We include these columns to help with tracking the progress we make with \ModelOne. We also report performance of this model on our Round~2 dataset (described below in \secref{sec:round2}), again to help with tracking progress and understanding the two rounds.

\subsection{Harvesting Sentences}\label{sec:naturalistic-data}

Our first round of data collection focused on finding naturally occurring sentences that would challenge our \ModelZero. To do this, we harvested sentences from the Yelp Academic Dataset, using the version of the dataset that contains 8,021,122 reviews.

The sampling process was designed so that 50\% of the sentences fell into two groups: those that occurred in 1-star reviews but were predicted by \ModelZero\ to be Positive, and those that occurred in 5-star reviews but were predicted by \ModelZero\ to be Negative. The intuition here is that these would likely be examples that fooled our model. Of course, negative reviews can (and often do) contain positive sentences, and vice-versa. This motivates the validation stage that we describe next.

\begin{table*}[tp]
  \centering
  \begin{tabular}{p{0.5\linewidth} c c}
  \toprule
  Sentence & \ModelZero & Responses \\
  \midrule
  Good food nasty attitude by hostesses .          & neg &  \textbf{mix, mix, mix}, neg, neg \\
  Not much of a cocktail menu that I saw.          & neg &  \textbf{neg, neg, neg, neg, neg} \\
  I scheduled the work for 3 weeks later.          & neg &  \textbf{neu, neu, neu, neu}, pos \\
  I was very mistaken, it was much more!           & neg &  neg, \textbf{pos, pos, pos, pos} \\[1ex]
  It is a gimmick, but when in Rome, I get it.     & neu &  \textbf{mix, mix, mix}, neu, neu \\
  Probably a little pricey for lunch.              & neu &  mix, \textbf{neg, neg, neg, neg} \\
  But this is strictly just my opinion.            & neu &  \textbf{neu, neu, neu, neu}, pos \\
  The price was okay, not too pricey.              & neu &  mix, neu, \textbf{pos, pos, pos} \\[1ex]
  The only downside was service was a little slow. & pos &  \textbf{mix, mix, mix}, neg, neg \\
  However there is a 2 hr seating time limit.      & pos &  mix, \textbf{neg, neg, neg}, neu \\
  With Alex, I never got that feeling.             & pos &  \textbf{neu, neu, neu, neu}, pos \\
  Its ran very well by management.                 & pos &  \textbf{pos, pos, pos, pos, pos} \\
\bottomrule
  \end{tabular}
  \caption{Round~1 train set examples, randomly selected from each combination of \ModelZero\ prediction and majority label, but limited to examples with 30--50 characters. \Appref{app:random-examples} provides fully randomly selected examples.}
  \label{tab:round1-sample}
\end{table*}

\subsection{Validation}\label{sec:round1-validation}

Our validation task was conducted on Amazon Mechanical Turk. Workers were shown ten sentences and asked to label them according to the following four categories, which we give here with the glosses that were given to workers in instructions and in the annotation interface:
\begin{description}\setlength{\itemsep}{0pt}
\item[Positive:] The sentence conveys information about the author's positive evaluative sentiment.
\item[Negative:] The sentence conveys information about the author's negative evaluative sentiment.
\item[No sentiment:] The sentence does not convey anything about the author's positive or negative sentiment.
\item[Mixed sentiment:] The sentence conveys a mix of positive and negative sentiment with no clear overall sentiment.
\end{description}

We henceforth refer to the `No sentiment' category as `Neutral' to remain succinct and align with standard usage in the literature.

For this round, 1,978 workers participated in the validation process.  In the final version of the corpus, each sentence is validated by five different workers. To obtain these ratings, we employed an iterative strategy. Sentences were uploaded in batches of 3--5K and, after each round, we measured each worker's rate of agreement with the majority. We then removed from the potential pool those workers who disagreed more than 80\% of the time with their co-annotators, using a method of `unqualifying' workers that does not involving rejecting their work or blocking them \citep{MTurk:SoftBan:2017}. We then obtained additional labels for examples that those `unqualified' workers annotated. Thus, many examples received more than five responses over the course of validation. The final version of \ASenti\ keeps only the responses from the highest-rated workers. This led to substantial increase in dataset quality by removing a lot of labels that seemed to us to be randomly assigned. \Appref{app:validation} describes the process in more detail, and our Datasheet enumerates the known unwanted biases that this process can introduce.

\begin{table}[tp]
  \centering
  \newcommand{\spacer}{\hspace{12pt}}
  \setlength{\tabcolsep}{3pt}
    \begin{tabular}[c]{l r @{\spacer} *{3}{r} }
      \toprule
                & \multicolumn{1}{c@{\spacer}}{Dist} 
                & \multicolumn{3}{c}{Majority Label} \\
                & \multicolumn{1}{c@{\spacer}}{Train}         
                & Train & Dev & Test \\
      \midrule
      Positive    & 130,045 & 21,391 & 1,200 & 1,200 \\
      Negative    & 86,486  & 14,021 & 1,200 & 1,200 \\
      Neutral     & 215,935 & 45,076 & 1,200 & 1,200 \\
      Mixed       & 39,829  &  3,900 &     0 &     0 \\
      No Majority & --      & 10,071 &     0 &     0 \\[1ex]
      Total       & 472,295 & 94,459 & 3,600 & 3,600 \\
      \bottomrule
    \end{tabular}
  \caption{Round~1 Dataset. `Dist' refers to the individual responses (five per example). We omit the comparable Dev and Test numbers to save space. The Majority Label is the one chosen by at least three of the five workers, if there is such label. For Majority Label Train, the Positive, Negative, and Neutral categories contain a total of 80,488 examples.}
  \label{tab:round1-data}
\end{table}

\begin{table*}[tp]
  \centering 
  \begin{subtable}[b]{1\textwidth}
  \setlength{\tabcolsep}{14pt}
  \centering
    \begin{tabular}[b]{l *{6}{r} }
      \toprule
      & CR    & IMDB   & SST-3 & Yelp & Amazon & Round~1 \\
      \midrule
      Positive  & 2,405 & 12,500 & 128,016  & 29,841 & 133,411 & 339,748\\
      Negative  & 1,366 & 12,500 & 104,832  & 30,086 & 133,267 & 252,630\\
      Neutral   & 0     & 0      & 244,974  & 30,073 & 133,322 & 431,870 \\[1ex]
      Total     & 3,771 & 25,000 & 477,822 & 90,000 & 400,000 & 1,024,248\\
      \bottomrule
    \end{tabular}
    \caption{\ModelOne\ training data. CR and IMDB are unchanged from \tabref{tab:model0-train-data}. SST-3 is processed the same way as before, but we repeat the dataset 3 times to give it more weight. For Yelp and Amazon, we include only 1-star, 3-star, and 5-star reviews, and we subsample from those categories, with the goal of down-weighting them overall and removing ambiguous reviews. Round~1 here uses distributional labels and is copied twice.}
    \label{tab:model1-train-data}
  \end{subtable}

  \vspace{6pt}
  
  \begin{subtable}[b]{1\textwidth}
    \centering
    \newcommand{\spacer}{\hspace{28pt}}
    \begin{tabular}[b]{l *{2}{r} @{\spacer} *{2}{r} @{\spacer} *{2}{r} @{\spacer} *{2}{r} @{\spacer} *{2}{r} }
      \toprule
      & \multicolumn{2}{c@{\spacer}}{SST-3} 
      & \multicolumn{2}{c@{\spacer}}{Yelp} 
      & \multicolumn{2}{c@{\spacer}}{Amazon} 
      & \multicolumn{2}{c@{\spacer}}{Round~1} 
      & \multicolumn{2}{c}{Round~2}\\
      & Dev & Test & Dev & Test  & Dev & Test & Dev & Test & Dev & Test\\
      \midrule
      Positive       & 84.6 & 88.6 & 80.0 & 83.1 & 83.3 & 83.3 & 81.0 & 80.4 & 33.3 & 33.3\\
      Negative       & 82.7 & 84.4 & 79.5 & 79.6 & 78.7 & 78.8 & 80.5 & 80.2 & 33.3 & 33.3\\
      Neutral       & 40.0 & 45.2 & 56.7 & 56.6 & 55.5 & 55.4 & 83.1 & 83.5 & 33.3 & 33.3\\[1ex]
      Macro avg & 69.1 & 72.7 & 72.1 & 73.1 & 72.5 & 72.5 & 81.5 & 81.4 & 33.3 & 33.3 \\
      \bottomrule
    \end{tabular}
    \caption{\ModelOne\ performance (F1 scores) on external assessment
      datasets (\tabref{tab:external-assess-data}), as well as our Round~1 and Round~2 datasets.
      Chance performance for this model on Round~2 is by design (\secref{sec:round2-data}).}
    \label{tab:model1-assess}
  \end{subtable}
  \caption{\ModelOne\ summary.}
  \label{tab:model1}
\end{table*}

\subsection{Round 1 Dataset}\label{sec:round1-dataset}

The resulting Round~1 dataset is summarized in \tabref{tab:round1-data}. \Tabref{tab:round1-sample} provides train-set examples for every combination of \ModelZero\ prediction and majority validation label. The examples were randomly sampled with a restriction to 30--50 characters. \Appref{app:random-examples} provides examples that were truly randomly selected.

Because each sentence has five ratings, there are two perspectives we can take on the dataset:

\paragraph{Distributional Labels} We can repeat each example with each of its labels. For instance, the first sentence in \tabref{tab:round1-sample} would be repeated three times with `Mixed' as the label and twice with `Negative'. For many classifier models, this reduces to labeling each example with its probability distribution over the labels. This is an appealing approach to creating training data, since it allows us to make use of all the examples,\footnote{For `Mixed' labels, we create two copies of the example, one labeled `Positive', the other `Negative'.} even those that do not have a majority label, and it allows us to make maximal use of the labeling information. In our experiments, we found that training on the distributional labels consistently led to slightly better models.

\paragraph{Majority Label} We can take a more traditional route and infer a label based on the distribution of labels. In \tabref{tab:round1-data}, we show the labels inferred by assuming that an example has a label just in case at least three of the five annotators chose that label. This is a conservative approach that creates a fairly large `No Majority' category.  More sophisticated approaches might allow us to make fuller use of the examples and account for biases relating to annotator quality and example complexity (see \secref{sec:crowdsourcing}). We set these options aside for now because our validation process placed more weight on the best workers we could recruit (\secref{sec:round1-validation}).

The Majority Label splits given by \tabref{tab:round1-data} are designed to ensure five properties: (1) the classes are balanced, (2) \ModelZero\ performs at chance, (3) the review-level rating associated with the sentence has no predictive value, (3) at least four of the five workers agreed, and (5) the majority label is Positive, Negative, or Neutral. (This excludes examples that received a Mixed majority and examples without a majority label at all.)

Over the entire round, 47\% of cases are such that the validation majority label is Positive, Negative, or Neutral and \ModelZero\ predicted a different label.

\subsection{Estimating Human Performance}\label{sec:round1-human}

\Tabref{tab:round1-agr} provides a conservative estimate of human F1 in order to have a quantity that is comparable to our model assessment metrics. To do this, we randomize the responses for each example to create five synthetic annotators, and we calculate the precision, recall, and F1 scores for each of these annotators with respect to the gold label. We average those scores. This is a conservative estimate of human performance, since it heavily weights the single annotator who disagreed for the cases with 4/5 majorities. We can balance this against the fact that 614 workers (out of 1,280) \emph{never} disagreed with the majority label (see \appref{app:validation} for the full distribution of agreement rates). However, it seems reasonable to say that a model has solved the round if it achieves comparable scores to our aggregate F1 -- a helpful signal to start a new round.

\begin{table}[tp]
  \centering
    \begin{tabular}[c]{l r r }
      \toprule               
                & Dev & Test \\
      \midrule
      Positive  & 88.1 & 87.8 \\
      Negative  & 89.2 & 89.3 \\
      Neutral   & 86.6 & 86.9 \\[1ex]
      Macro avg & 88.0 & 88.0 \\
      \bottomrule
    \end{tabular}
  \caption{Estimates of human performance (F1 scores) on the Round~1 dataset. The estimates come from comparing random synthesized human annotators against the gold labels using the response distributions in the dataset. The Fleiss Kappas for the dev and tests set are 0.616 and 0.615, respectively. We offer F1s as a way of tracking model performance to determine when the round is ``solved'' and a new round of data should be collected. However, we note that 614 of our 1,280 workers never disagreed with the gold label.}
  \label{tab:round1-agr}
\end{table}

\section{Round 2: Dynabench}\label{sec:round2}

In Round 2, we leverage the Dynabench platform to begin creating a new dynamic benchmark for sentiment analysis. Dynabench is an open-source application that runs in a browser and facilitates dataset collection efforts in which workers seek to construct examples that fool a model (or ensemble of models) but make sense to other humans.

\subsection{\ModelOne}\label{sec:model1}

\ModelOne\ was created using the same general methods as for \ModelZero\ (\secref{sec:model0}): we begin with RoBERTa parameters and add a three-way sentiment classifier head. The differences between the two models lie in the data they were trained on.

\Tabref{tab:model1-train-data} summarizes the training data for \ModelOne. In general, it uses the same datasets as we used for \ModelZero, but with a few crucial changes. First, we subsample the large Yelp and Amazon datasets to ensure that they do not dominate the dataset, and we include only 1-star, 3-star, and 5-star reviews to try to reduce the number of ambiguous examples. Second, we upsample SST-3 by a factor of 3 and our own dataset by a factor of 2, using the distributional labels for our dataset (\secref{sec:round1-dataset}). This gives roughly equal weight, by example, to our dataset as to all the others combined. This makes sense given our general goal of doing well on our dataset and, especially, of shifting the nature of the Neutral category to something more semantically coherent than what the other corpora provide.

\Tabref{tab:model1-assess} summarizes the performance of our model on the same evaluation sets as are reported in \tabref{tab:model1-assess} for \ModelZero. Overall, we see a small performance drop on the external datasets, but a huge jump in performance on our dataset (Round~1). While it is unfortunate to see a decline in performance on the external datasets, this is expected if we are shifting the label distribution with our new dataset -- it might be an inevitable consequence of hill-climbing in our intended direction.

\subsection{Dynabench Interface}

\Appref{app:dynabench} provides the Dynabench interface we created for \ASenti\ as well the complete instructions and training items given to workers. The essence of the task is that the worker chooses a label $y$ to target and then seeks to write an example that the model (currently, \ModelOne) assigns a label other than $y$ but that other humans would label $y$. Workers can try repeatedly to fool the model, and they get feedback on the model's predictions as a guide for how to fool it.

\subsection{Methods}\label{sec:dynabench-data}

We consider two conditions:
\begin{description}\setlength{\itemsep}{0pt}
\item[Prompt] Workers are shown a sentence and given the opportunity to modify it as part of achieving their goal. Prompts are sampled from parts of the Yelp Academic Dataset not used for Round~1.

\item[No Prompt] Workers wrote sentences from scratch, with no guidance beyond their goal of fooling the model.
\end{description}
We piloted both versions and compared the results. Our analyses are summarized in \secref{sec:prompts}. The findings led us to drop the No Prompt condition and use the Prompt condition exclusively, as it clearly leads to examples that are more naturalistic and linguistically diverse.

For Round~2, our intention was for each prompt to be used only once, but prompts were repeated in a small number of cases. We have ensured that our dev and test sets contain only sentences derived from unique prompts (\secref{sec:round2-data}).

\subsection{Validation}\label{sec:round2-validation}

We used the identical validation process as described in \secref{sec:round1-validation}, getting five responses for each example as before. This again opens up the possibility of using label distributions or inferring individual labels. 395 workers participated in the validation process for Round~2. See \appref{app:validation} for additional details.

\begin{table}[htp]
    \centering
    \newcommand{\spacer}{\hspace{12pt}}
    \begin{tabular}[c]{l r @{\spacer} *{3}{r} }
      \toprule
      & \multicolumn{1}{c@{\spacer}}{Dist} 
      & \multicolumn{3}{c}{Majority Label} \\
      & \multicolumn{1}{c@{\spacer}}{Train}
      & Train & Dev & Test \\
      \midrule
      Positive    & 32,551 &    6,038 &  240 &  240 \\
      Negative    & 24,994  &   4,579 &  240 &  240\\
      Neutral     & 16,365  &   2,448 &  240 &  240 \\
      Mixed       & 18,765  &   3,334 &    0 &    0 \\
      No Majority & --      &   2,136 &    0 &    0 \\[1ex]
      Total       & 92,675  &  18,535 &  720 & 720\\
      \bottomrule
    \end{tabular}
    \caption{Round~2 splits using the framework described in \secref{sec:round2-data} and the criteria specified in \secref{sec:round2-data}.}
  \label{tab:round2-data}
\end{table}

\subsection{Round 2 Dataset}\label{sec:round2-data}

\Tabref{tab:round2-data} summarizes our Round~2 dataset. Overall, workers' success rate in fooling \ModelOne\ is about 19\%, which is much lower than the comparable value for Round~1 (47\%). There seem to be at least three central reasons for this. First, \ModelOne\ is hard to fool, so many workers reach the maximum number of attempts. We retain the examples they enter, as many of them are interesting in their own right. Second, some workers seem to get confused about the true goal and enter sentences that the model in fact handles correctly. Some non-trivial rate of confusion here seems inevitable given the cognitive demands of the task, but we have taken steps to improve the interface to minimize this factor. Third, a common strategy is to create examples with mixed sentiment; the model does not predict this label, but it is chosen at a high rate in validation.

Despite these complicating factors, we can construct splits that meet our core goals: (1) \ModelOne\ performs at chance on the dev and test sets, and (2) the dev and test sets contain only examples where the majority label was chosen by at least four of the five workers. In addition, (3) our dev and test sets contain only examples from the Prompt condition (the No Prompt cases are in the train set, and flagged as such), and (4) all the dev and test sentences are derived from unique prompts to avoid leakage between train and assessment sets and reduce unwanted correlations within the assessment sets. \Tabref{tab:round2-data} summarizes these splits.

\Tabref{tab:round2-sample} provides train examples from Round~2 sampled using the same criteria we used for \tabref{tab:round1-sample}: the examples are randomly chosen to show every combination of model prediction and majority label, with the restriction that the examples be 30--50 characters long. (\Appref{app:random-examples} gives fully randomly selected examples.)

\begin{table*}[tp]
  \centering
  \begin{tabular}{p{0.5\linewidth} c c}
  \toprule
    Sentence & \ModelOne & Responses \\
    \midrule
    The place was somewhat good and not well          &  neg &  \textbf{mix, mix, mix, mix}, neg \\
    I bought a new car and met with an accident.      &  neg &  \textbf{neg, neg, neg, neg, neg} \\
    The retail store is closed for now at least.      &  neg &  \textbf{neu, neu, neu, neu, neu} \\
    Prices are basically like garage sale prices.     &  neg &  neg, neu, \textbf{pos, pos, pos} \\[1ex]
    That book was good.  I need to get rid of it.     &  neu &  \textbf{mix, mix, mix}, neg, pos \\
    I REALLY wanted to like this place                &  neu &  mix, \textbf{neg, neg, neg}, pos \\
    But I'm going to leave my money for the next vet. &  neu &  neg, \textbf{neu, neu, neu, neu} \\
    once upon a time the model made a super decision. &  neu &  \textbf{pos, pos, pos, pos, pos} \\[1ex]
    I cook my caribbean food and it was okay          &  pos & \textbf{ mix, mix, mix}, pos, pos \\
    This concept is really cool in name only.         &  pos &  mix, \textbf{neg, neg, neg}, neu \\
    Wow, it'd be super cool if you could join us      &  pos &  \textbf{neu, neu, neu, neu}, pos \\
    Knife cut thru it like butter! It was great.      &  pos &  \textbf{pos, pos, pos, pos, pos} \\
    \bottomrule
  \end{tabular}
  \caption{Round~2 train set examples, randomly selected from each combination of \ModelOne\ prediction and majority label, but limited to examples with 30--50 characters. \Appref{app:random-examples} provides fully randomly selected examples.}
  \label{tab:round2-sample}
\end{table*}

\subsection{Estimating Human Performance}\label{sec:round2-human}

\Tabref{tab:round2-agr} provides estimates of human F1 for Round~2 using the same methods as described in \secref{sec:round1-human} and given in the corresponding table for Round~1 (\tabref{tab:round1-agr}). We again emphasize that these are very conservative estimates. We once again had a large percentage of workers (116~of~244) who never disagreed with the gold label on the examples they rated, suggesting that human performance can approach perfection. Nonetheless, the estimates we give here seem useful for helping us decide whether to continue hill-climbing on this round or begin creating new rounds.

\begin{table}[tp]
  \centering
    \begin{tabular}[c]{l r r }
      \toprule               
                & Dev & Test \\
      \midrule
      Positive  & 91.0 & 90.9 \\
      Negative  & 91.2 & 91.0  \\
      Neutral   & 88.9 & 88.2  \\[1ex]
      Macro avg & 90.4 & 90.0  \\
      \bottomrule
    \end{tabular}
  \caption{Estimates of human performance (F1 scores) on the Round~2 dataset using the procedure described in \secref{sec:round1-human}. The Fleiss Kappas for the dev and tests set are 0.681 and 0.667, respectively. The F1s are conservative estimates; 116 of our 244 workers never disagreed with the gold label for this round..}
  \label{tab:round2-agr}
\end{table}

\section{Discussion}

We now address a range of issues that our methods raise but that we have so far deferred in the interest of succinctly reporting on the methods themselves.

\subsection{The Role of Prompts}\label{sec:prompts}

As discussed in \secref{sec:round2}, we explored two methods for collecting original sentences on Dynabench: with and without a prompt sentence that workers could edit to achieve their goal. We did small pilot rounds in each condition and assessed the results. This led us to use the Prompt condition exclusively. This section explains our reasoning more fully.

\begin{figure*}[tp]
  \centering
  \begin{subfigure}[t]{0.32\linewidth}
    \centering
    \includegraphics[width=1\linewidth]{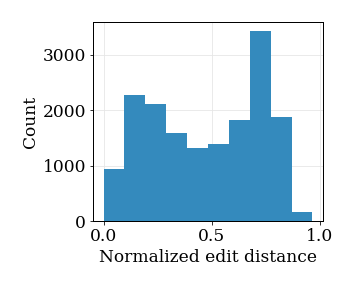}
    \caption{Normalized edit distances between the prompt and the example.}
    \label{fig:edit-distance}
  \end{subfigure}
  \hfill
  \begin{subfigure}[t]{0.32\linewidth}
    \centering
    \includegraphics[width=1\linewidth]{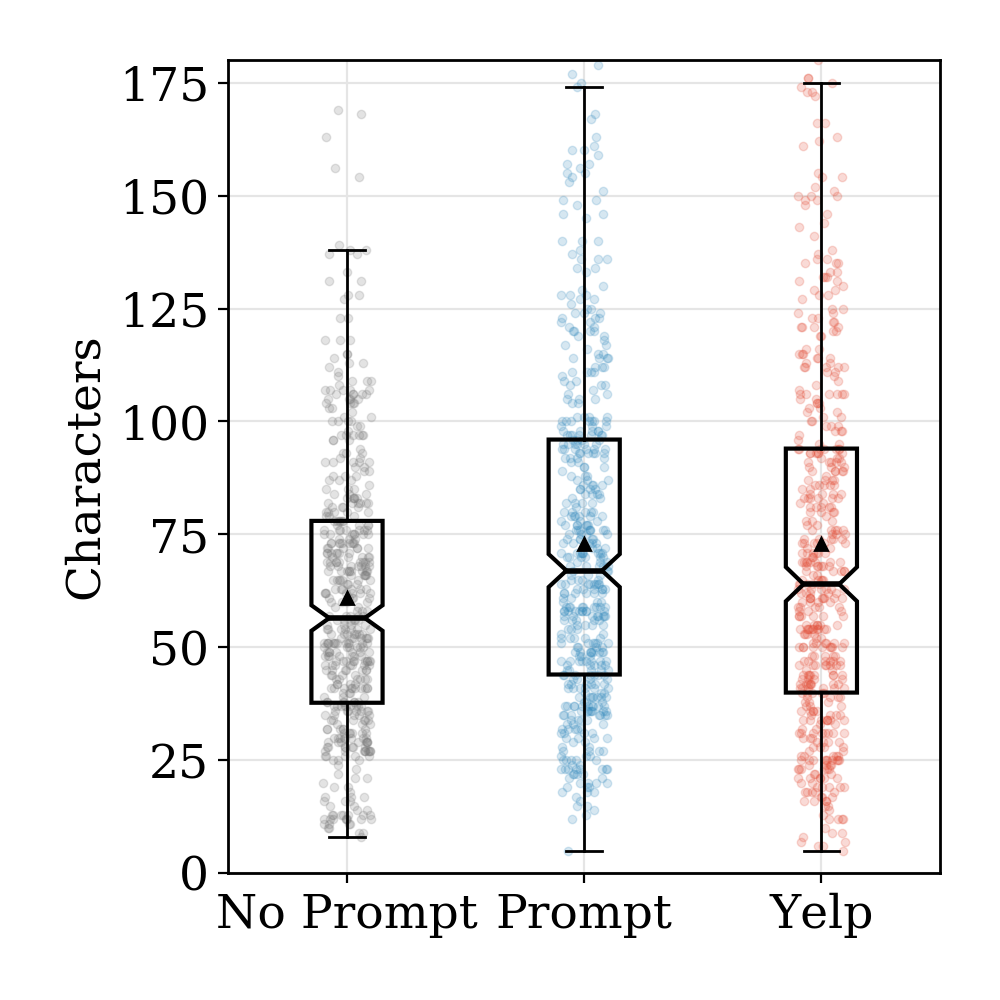}
    \caption{String lengths. The picture is essentially the same for tokenized word counts.}
    \label{fig:string-lengths}
  \end{subfigure}
  \hfill
  \begin{subfigure}[t]{0.32\linewidth}
    \centering
    \includegraphics[width=1\linewidth]{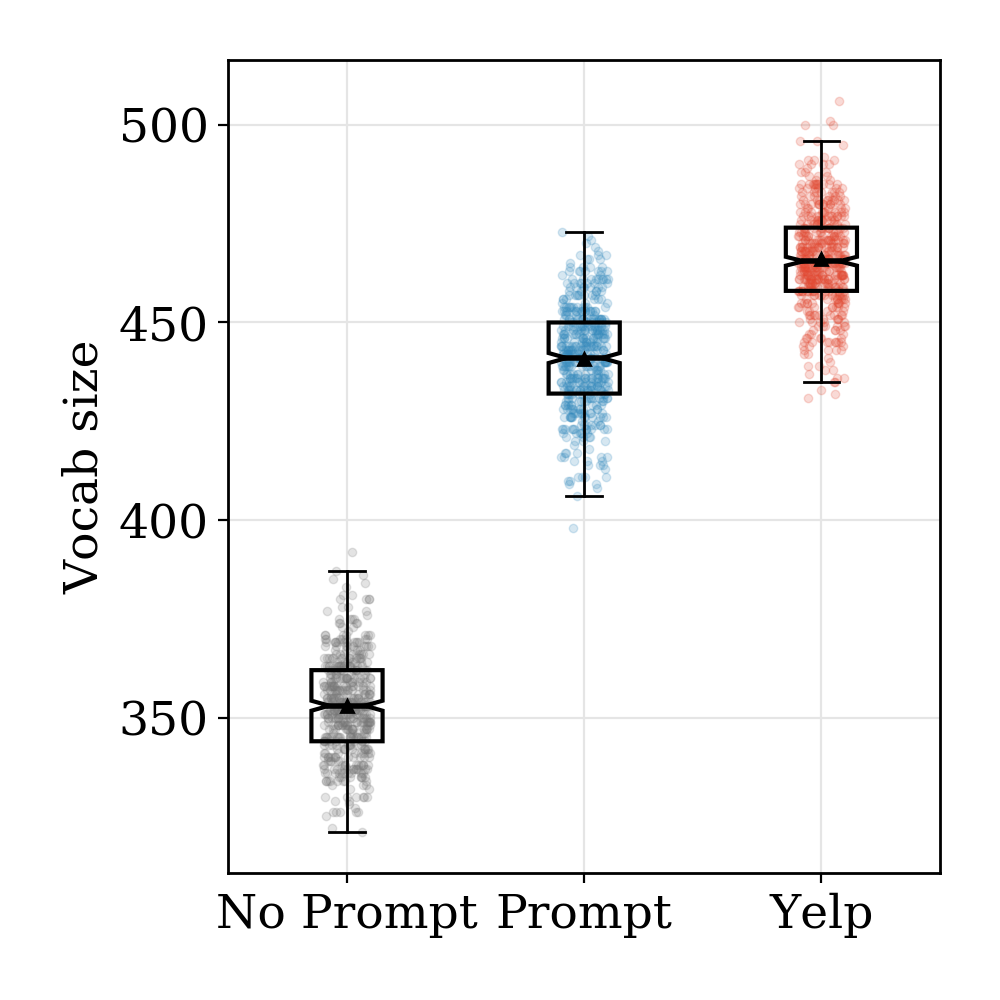}
    \caption{Vocabulary sizes in samples of 100 examples (500 samples with replacement).}
    \label{fig:vocabs}
  \end{subfigure}
  \caption{The `Prompt' and `No Prompt' conditions.}
  \label{fig:prompts}
\end{figure*}

First, we note that workers did in fact make use of the prompts. In \figref{fig:edit-distance}, we plot the Levenshtein edit distance between the prompts provided to annotators and the examples the annotators produced, normalized by the length of the prompt or the example, whichever is longer. There is a roughly bimodal distribution in this plot, where the peak on the right represents examples generated by the annotator tweaking the prompt slightly and the peak on the left represents examples where they deviated significantly from the prompt. Essentially no examples fall at the extreme ends (literal reuse of the prompt; complete disregard for the prompt).

Second, we observe that examples generated in the Prompt condition are generally longer than those in the No Prompt condition, and more like our Round~1 examples. \Figref{fig:string-lengths} summarizes for string lengths; the picture is essentially the same for tokenized word counts. In addition, the Prompt examples have a more diverse vocabulary overall. \Figref{fig:vocabs} provides evidence for this: we sampled 100 examples from each condition 500 times, sampled five words from each example, and calculated the vocabulary size (unique token count) for each sample. (These measures are intended to control for the known correlation between token counts and vocabulary sizes; \citealt{Baayen01WORD}.) The Prompt-condition vocabularies are much larger, and again more similar to our Round~1 examples.

Third, a qualitative analysis further substantiates the above picture.  For example, many workers realized that they could fool the model by attributing a sentiment to another group and then denying it, as in ``They said it would be great, but they were wrong''. As a result, there are dozens of examples in the No Prompt condition that employ this strategy. Individual workers hit upon more idiosyncratic strategies and repeatedly used them. This is just the sort of behavior that we know can create persistent dataset artifacts. For this reason, we include No Prompt examples in the training data only, and we make it easy to identify them in case one wants to handle them specially.

\subsection{The Neutral Category}\label{sec:neutral}

For both \ModelZero\ and \ModelOne, there is consistently a large gap between performance on the Neutral category and performance on the other categories, but only for the external datasets we use for evaluation. For our dataset, performance across all three categories is fairly consistent. We hypothesized that this traces to semantic diversity in the Neutral categories for these external datasets. In review corpora, three-star reviews can signal neutrality, but they are also likely to signal mixed sentiment or uncertain overall assessments. Similarly, where the ratings are assigned by readers, as in the SST, it seems likely that the middle of the scale will also be used to register mixed and uncertain sentiment, along with a real lack of sentiment.

To further support this hypothesis, we ran the SST dev set through our validation pipeline. This leads to a completely relabeled dataset with five ratings for each example and a richer array of categories. \Tabref{tab:sst-validation} compares these new labels with the labels in SST-3. Overall, the two label sets are closely aligned for Positive and Negative. However, the SST-3 Neutral category has a large percentage of cases falling into Mixed and No Majority, and it is overall by far the least well aligned with our labeling of any of the categories. \Appref{app:sst-cmp} gives a random sample of cases where the two label sets differ with regard to the Neutral category. It also provides all seven cases of sentiment confusion. We think these comparisons favor our labels over SST's original labels.

\begin{table}[tp]
\centering
\begin{tabular}{lrrr}
\toprule
& \multicolumn{3}{c}{SST-3} \\
 &  Positive &    Negative &   Neutral \\
\midrule
Positive &  367 &    2 &  64 \\
Negative &    5 &  359 &  57 \\
Neutral  &   23 &    8 &  44 \\
Mixed    &   34 &   35 &  39 \\
No Majority   &   15 &   24 &  25 \\
\bottomrule
\end{tabular}
\caption{Comparison of the SST-3 labels (dev set) with labels derived from our separate validation of this dataset.}
\label{tab:sst-validation}
\end{table}

\subsection{Fine-Tuning}\label{sec:fine-tuning}

Our \ModelOne\ was trained from scratch (beginning with RoBERTa) parameters. An appealing alternative would be to begin with \ModelZero\ and fine-tune it on our Round~1 data. This would be more efficient, and it might naturally lead to the Round~1 data receiving the desired overall weight relative to the other datasets.  Unfortunately, our attempts to fine-tune in this way led to worse models, and the problems generally traced to very low performance on the Neutral category.

To study the effect of our dataset on \ModelOne\ performance, we employ the ``fine-tuning by inoculation'' method of  \citet{liu-etal-2019-inoculation}. We first divide our Round~1 train set into small subsets via random sampling. Then, we fine-tune our \ModelZero\ using these subsets of Round~1 train with non-distributional labels. We early-stop our fine-tuning process if performance on the Round~0 dev set of \ModelZero\ (SST-3 dev) has not improved for five epochs. Lastly, we measure model performance with Round~1 dev (SST-3 dev plus Round~1 dev) and our external evaluation sets (\tabref{tab:external-assess-data}).

\begin{figure*}[tp]
\minipage{0.98\textwidth}
  \includegraphics[width=\linewidth]{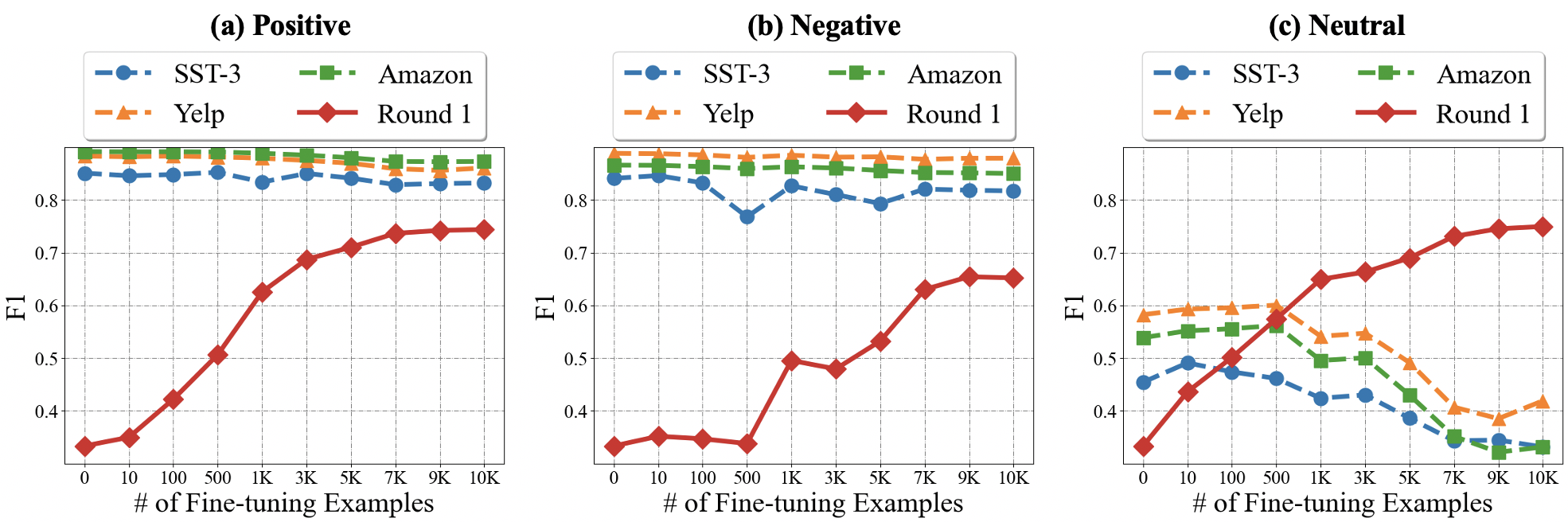}
\endminipage 
\caption{Inoculation by fine-tuning results with different number of fine-tuning examples. (a-c): F1 score on different development sets for three categories: Positive, Negative and Neutral. }
\label{fig:inoculation}
\end{figure*}

\Figref{fig:inoculation} presents F1 scores for our three class labels using this method. We observe that model performance on Round~1 dev increases for all three labels given more training examples. The F1 scores for the Positive and Negative classes remain high, but they begin to drop slightly  with larger samples. The F1 scores on SST-3 dev show larger perturbations. However, the most striking trends are for the Neutral category, where the F1 score on Round~1 dev increases steadily while the F1 scores on the three original development sets for \ModelZero\ decrease drastically. This is the general pattern that \citet{liu-etal-2019-inoculation} associate with dataset artifacts or label distribution shifts.

Our current  hypothesis is that the pattern we observe can be attributed, at least in large part, to label shift -- specifically, to the difference between our Neutral category and the other Neutral categories, as discussed in the preceding section. Our strategy of training from scratch seems less susceptible to these issues, though the label shift is still arguably a factor in the relatively poor performance we see on this category with our external validation sets.

\section{Conclusion}

We presented \ASenti, as the first stage in what we hope is an ongoing effort to create a dynamic benchmark for sentiment analysis that responds to scientific advances and concerns relating to real-world deployments of sentiment analysis models. \ASenti\ contains \totalExamples\ examples from two different sources: naturally occurring sentences and sentences created on the Dynabench Platform by workers who were actively trying to fool a strong sentiment model. All the sentences in the dataset are multiply-validated by crowdworkers in separate tasks. In addition, we argued for the use of prompt sentences on Dynabench to help workers avoid creative ruts and reduce the rate of biases and artifacts in the resulting dataset. We hope that the next step for \ASenti\ is that the community responds with models that solve both rounds of our task. That will in turn be our cue to launch another round of data collection to fool those models and push the field of sentiment forward by another step.

\section*{Acknowledgements}

Our thanks to the developers of the Dynabench Platform, and special thanks to our Amazon Mechanical Turk workers for their essential contributions to this project. This research is supported in part by faculty research grants from Facebook and Google.

\bibliography{anthology,adversarial-sentiment-bib}
\bibliographystyle{acl_natbib}

\newpage
\clearpage

\appendix

\section*{Appendix}

\section{\ModelZero}\label{app:model0}

To train our \ModelZero, we import weights from the pretrained RoBERTa-base model.\footnote{\url{https://dl.fbaipublicfiles.com/fairseq/models/roberta.base.tar.gz}} As in the original RoBERTa-base model \citep{liu2019roberta}, our model has 12 heads and 12 layers, with hidden layer size 768. The model uses byte-pair encoding as the tokenizer \citep{sennrich-etal-2016-neural}, with a maximum sequence length of 128. The initial learning rate is $2e{-5}$ for all trainable parameters, with a batch size of 8 per device (GPU). We fine-tuned for 3 epochs with a dropout probability of 0.1 for both attention weights and hidden states. To foster reproducibility, our training pipeline is adapted from the Hugging Face library \citep{Wolf2019HuggingFacesTS}.\footnote{\url{https://github.com/huggingface/transformers}} We used 6 $\times$ GeForce RTX 2080 Ti GPU each with 11GB memory. The training process takes about 15 hours to finish.

\section{Additional Details on Validation}\label{app:validation}

\subsection{Validation Interface}\label{app:validation-interface}

\begin{figure*}[tp]
  \centering
  \includegraphics[width=1\linewidth]{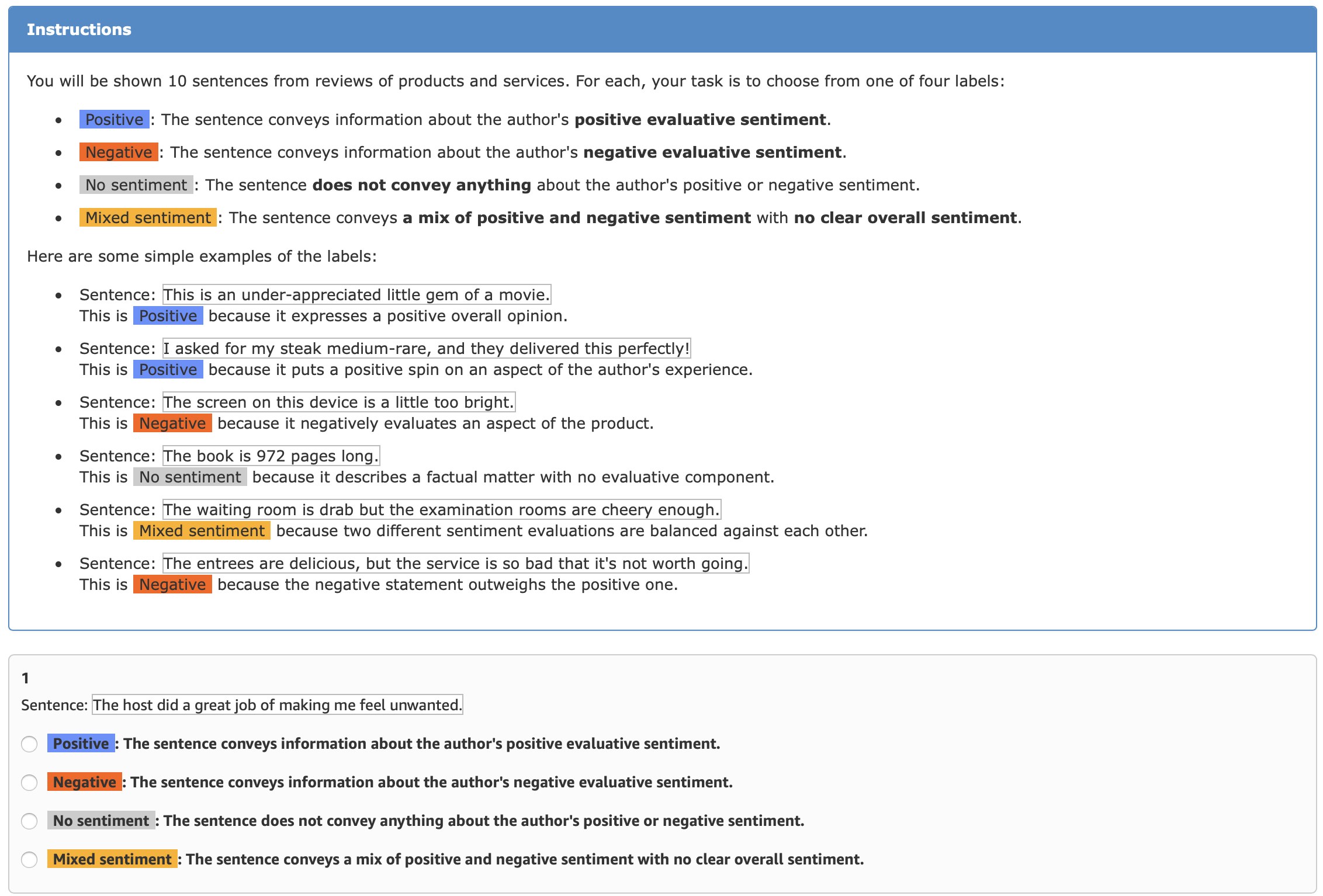}
  \caption{Validation interface.}
  \label{fig:validation-interface}
\end{figure*}

\Figref{fig:validation-interface} shows the interface for the validation task used for both Round~1 and Round~2. The top provides the instructions, and then one item is shown. The full task had ten items per Human Interface Task (HIT). Workers were paid US\$0.25 per HIT, and all workers were paid for all their work, regardless of whether we retained their labels.

\subsection{Worker Selection}

Examples were uploaded to Amazon's Mechanical Turk in batches of 3--5K examples. After each round, we assessed workers by the percentage of examples they labeled for which they agreed with the majority. For example, a worker who selects Negative where three of the other workers chose Positive disagrees with the majority for that example. If a worker disagreed with the majority more than 80\% of the time, we removed that worker from the annotator pool and revalidated the examples they labeled. This process was repeated iteratively over the course of the entire validation process for Round~1. Thus, many examples received more than 5 labels; we keep only those by the top-ranked workers according to agreement with the majority. We observed that this iterative process led to substantial improvements to the validation labels according to our own intuitions.

To remove workers from our pool, we used a method of `unqualifying', as described in \citep{MTurk:SoftBan:2017}. This method does no reputational damage to workers and is often used in situations where the requester must limit responses to one per worker (e.g., surveys). We do not know precisely why workers tend to disagree with the majority. The reasons are likely diverse. Possible causes include inattentiveness, poor reading comprehension, a lack of understanding of the task, and a genuinely different perspective on what examples convey. While we think our method mainly increased label quality, we recognize that it can introduce unwanted biases. We acknowledge this in our Datasheet, which is distributed with the dataset.

\subsection{Worker Distribution}

\begin{figure*}[htp]
  \begin{subfigure}{0.48\linewidth}
    \includegraphics[width=1\linewidth]{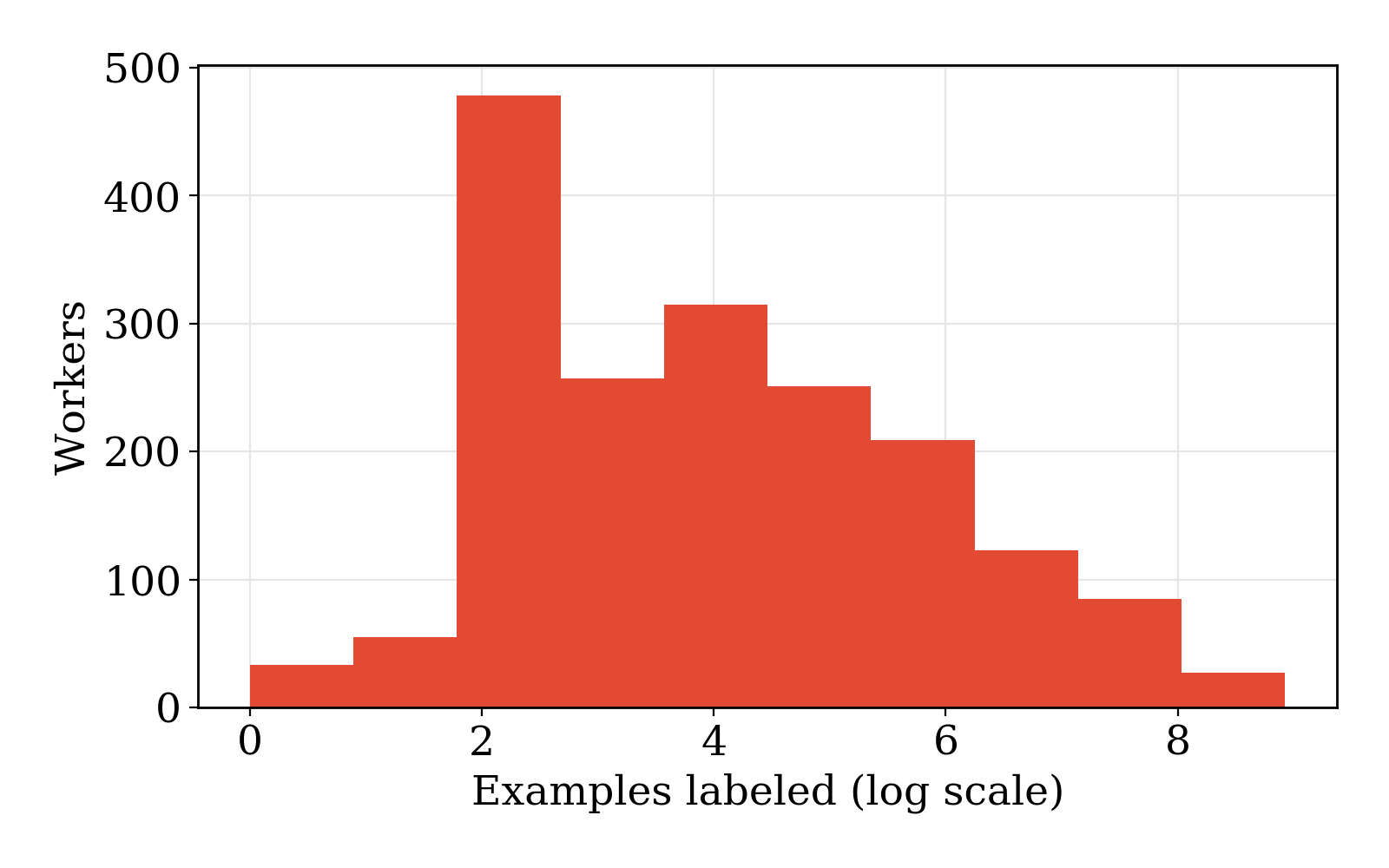}
    \caption{Round 1.}
  \end{subfigure}
  \hfill
  \begin{subfigure}{0.48\linewidth}
    \includegraphics[width=1\linewidth]{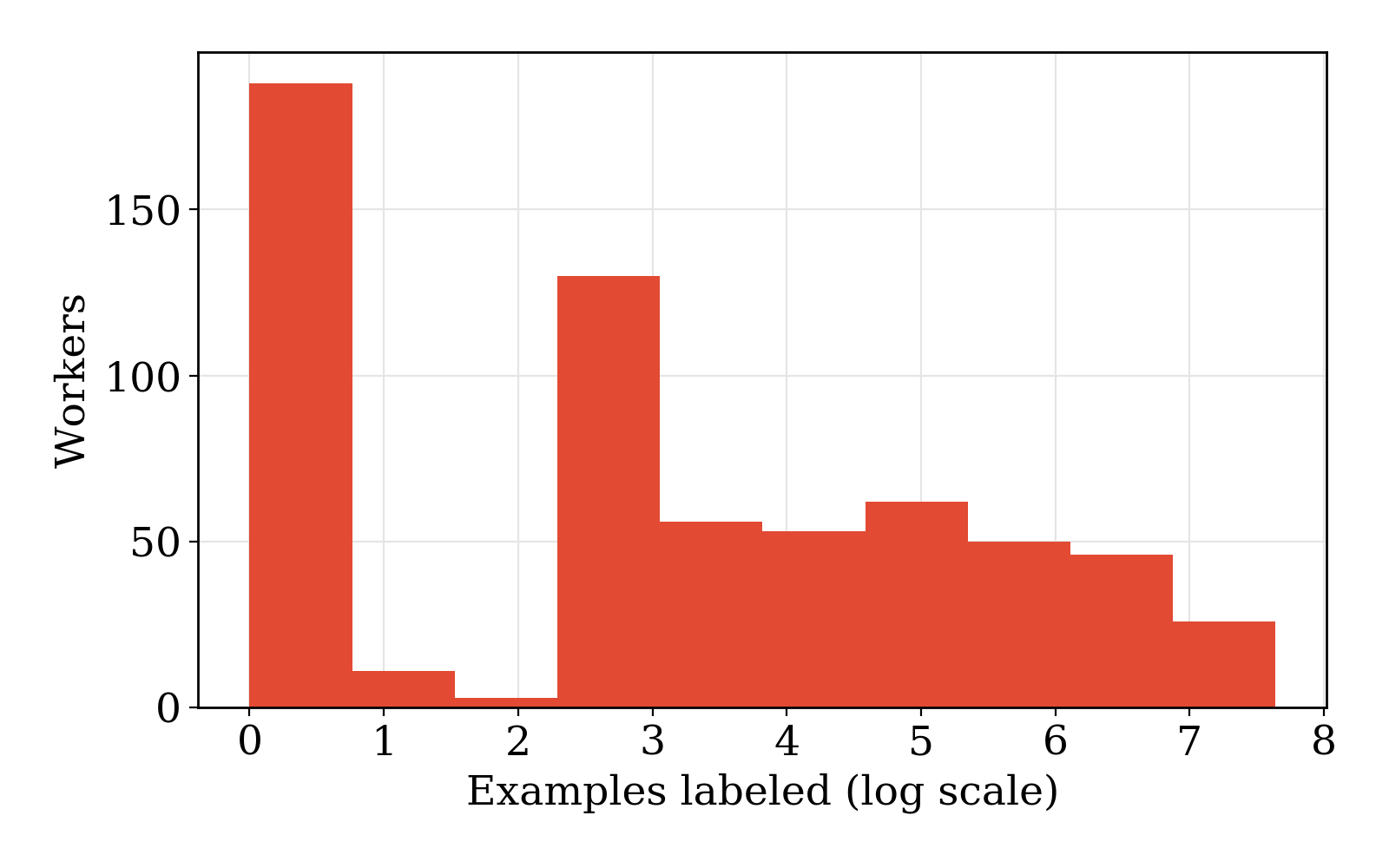}
    \caption{Round 2.}
  \end{subfigure}
  \caption{Worker distribution for the validation task.}
  \label{fig:workers}
\end{figure*}

\Figref{fig:workers} show the distribution of workers for the validation task for both rounds. In the final version of Round 1, the median number of examples per worker was 45 and the mode was 11. For Round~2, the median was 20 and the mode was 1.

\subsection{Worker Agreement with Gold Labels}\label{app:human-agr}

\begin{figure*}[tp]
  \begin{subfigure}{0.48\textwidth}
  \includegraphics[width=1\linewidth]{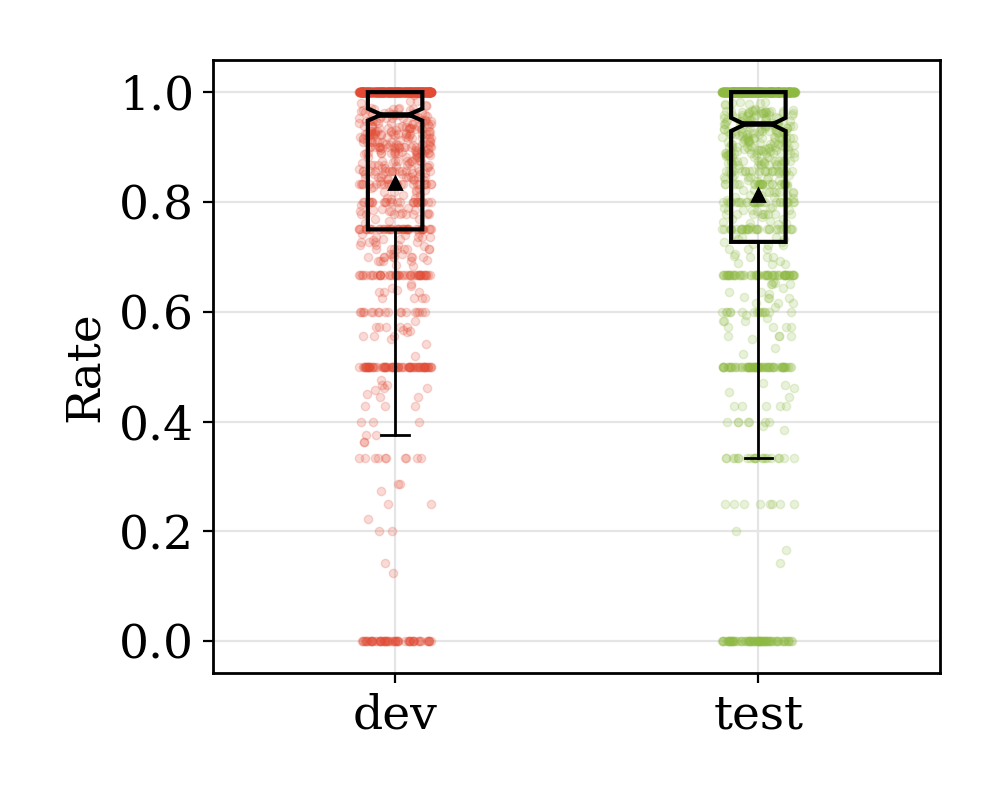}
  \caption{Round 1.}
  \label{fig:human-agr-round1-dist}
  \end{subfigure}
  \hfill\begin{subfigure}{0.48\textwidth}
  \includegraphics[width=1\linewidth]{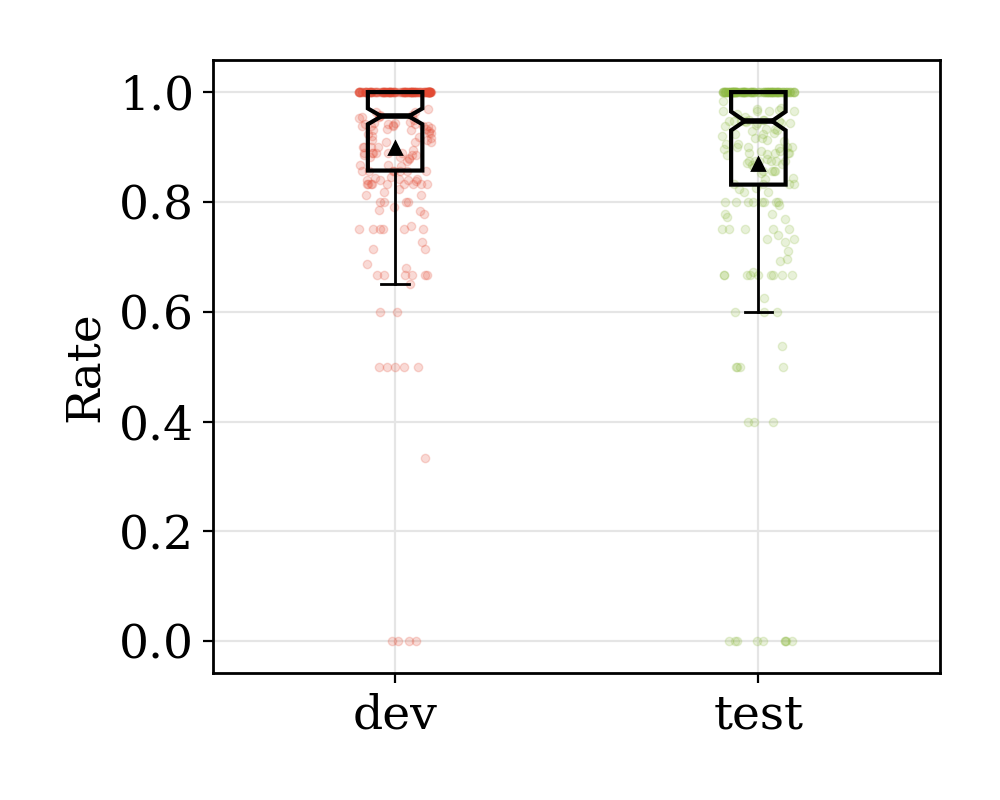}
  \caption{Round 2.}
  \label{fig:human-agr-round2-dist}
  \end{subfigure}  
\caption{Rates at which individual worker agree with the majority label. The y-axis gives, for each worker, the total number of examples for which they chose the majority label divided by the total number of cases they labeled over all.}
\label{fig:human-agr-dist}
\end{figure*}

\Figref{fig:human-agr-dist} summarizes the rates at which individual workers agree with the gold label. Across the dev and test sets for both rounds, substantial numbers of workers agreed with the gold label on all of the cases they labeled, and more than half were above 95\% for this agreement rate for both rounds.

\section{Additional Details on Dynabench Task}\label{app:dynabench}

\subsection{Interface for the Prompt Condition}

\Figref{fig:dynabench} shows an example of the Dynabench interface in the Prompt condition.

\subsection{Instructions}

\begin{figure*}[tp]
  \centering
  \includegraphics[width=1\linewidth]{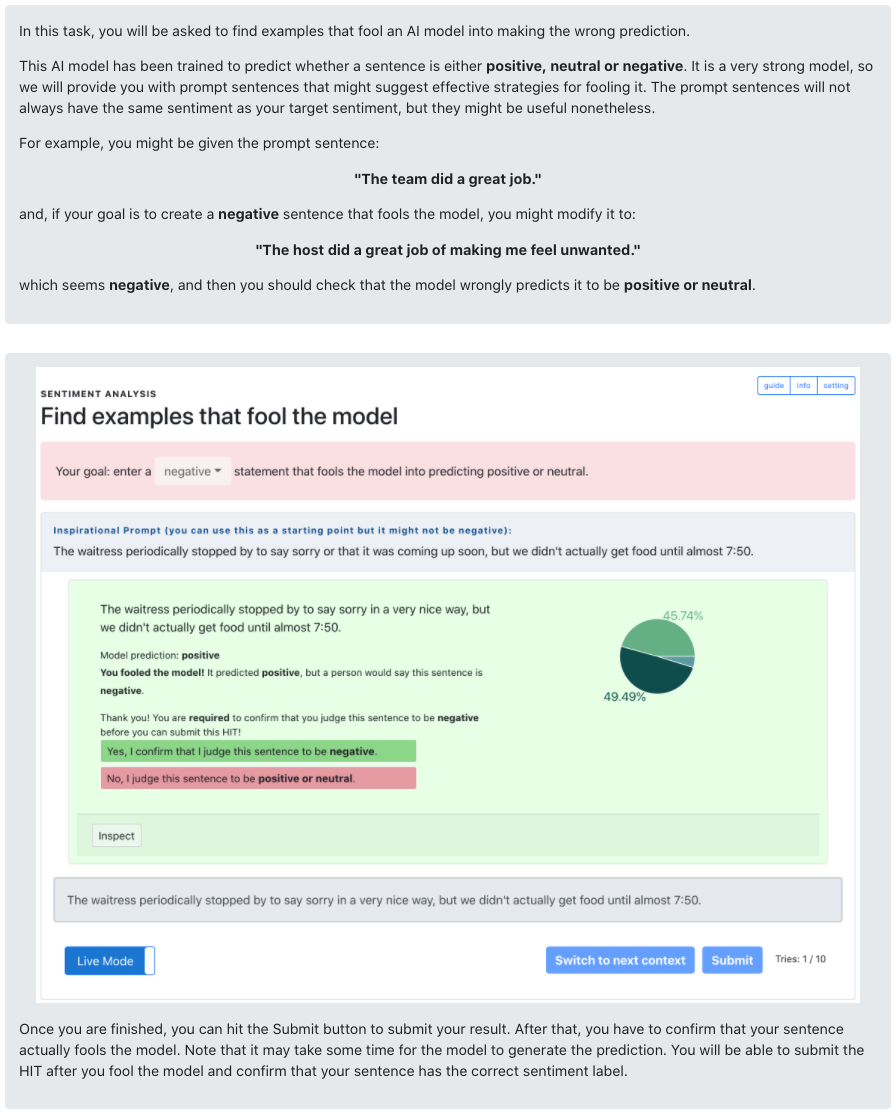}
  \caption{Dynabench instructions.}
  \label{fig:dynabench-instructions}
\end{figure*}

\Figref{fig:dynabench-instructions} provides the complete instructions for the Dynabench task, and \tabref{tab:onboarding-quiz} provides the list of comprehension questions we required workers to answer correctly before starting.

\subsection{Data Collection Pipeline}

For each task, a worker has ten attempts in total to find an example that fools the model. A worker can immediately claim their payments after submitting a single fooling example, or running out of attempts. The average number of attempts per task is two before the worker generates an example that they claim fools the model. Workers are paid US\$0.30 per task.

\begin{figure*}[tp]
  \centering
  \includegraphics[width=1\linewidth]{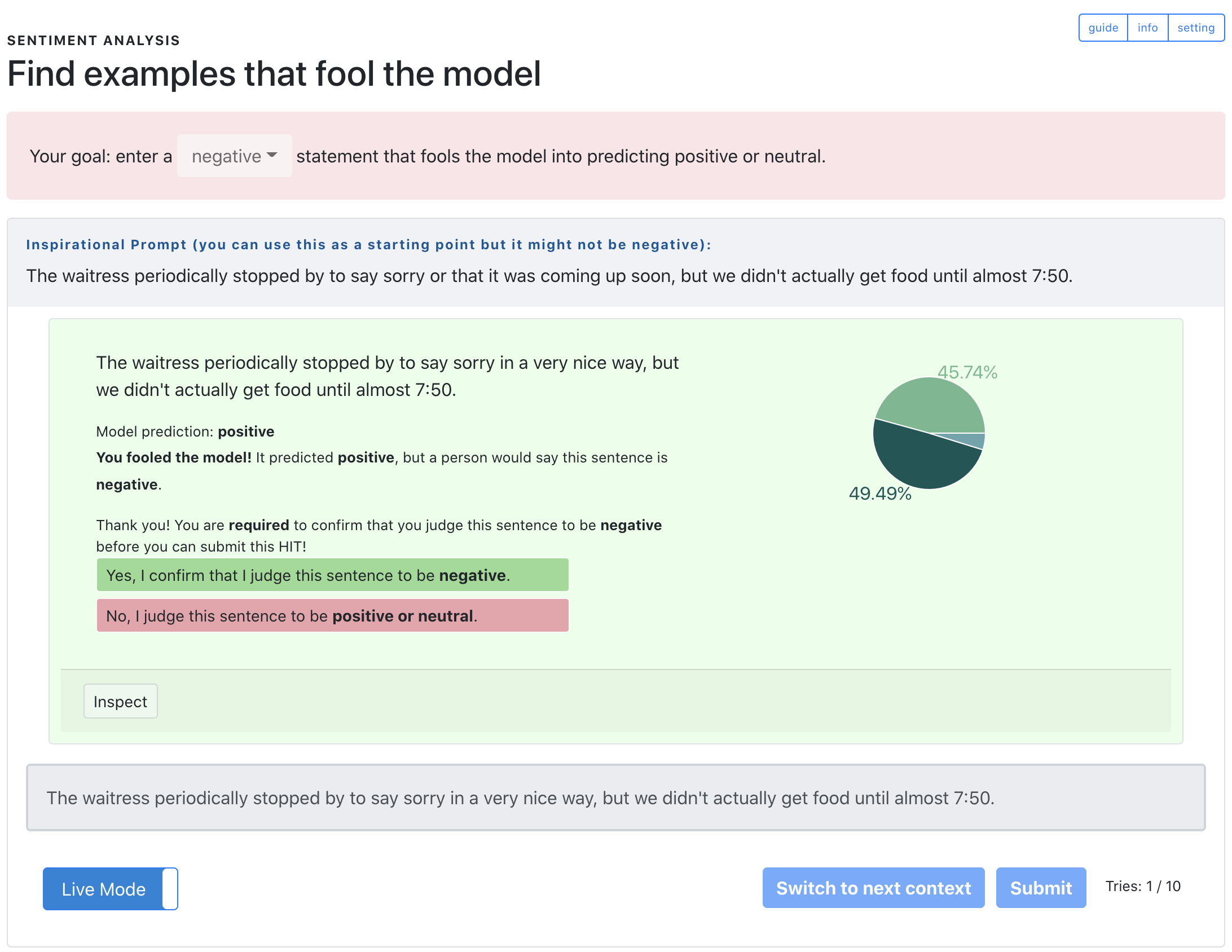}
  \caption{Dynabench interface.}
  \label{fig:dynabench}
\end{figure*}

A confirmation step is required if the model predicts incorrectly: we explicitly ask workers to confirm the examples they come up with are truly fooling examples. \Figref{fig:dynabench} shows this step in action.

To incentives workers, we pay a bonus of US\$0.30 for each truly fooling example according to our separate validation phase.

We temporarily disallow a worker to do our task if they fail to answer correctly to all our onboarding questions as in \Tabref{tab:onboarding-quiz} within five attempts. We also temporarily disallow a worker to do our task if they consistently cannot come up with truly fooling examples validated by us.

A worker must meet the following qualifications before accepting our tasks. First, a worker must reside in U.S.\ and speak English. Second, a worker must have completed at least 1,000 tasks on Amazon Mechanical Turk with an approval rating of 98\%. Lastly, a worker must not be in any of our temporarily disallowing worker pools.

We adapt the open-source software package Mephisto as our data collection tool for Amazon Mechanical Turk.\footnote{\url{https://github.com/facebookresearch/Mephisto}}

\section{SST-3 Validation Examples}\label{app:sst-cmp}

\Tabref{tab:sst-cmp} provides selected examples comparing the SST-3 labels with our revalidation of that dataset's labels.

\section{Randomly Selected Corpus Examples}\label{app:random-examples}

\Tabref{tab:random-examples} provides 10 truly randomly selected examples from each round's train set.

\begin{table*}[tp]
  \centering
  \renewcommand{\arraystretch}{1.2}
  \begin{tabular}{r@{ \ }p{0.77\linewidth} c}
  \toprule
    & Questions & Answers \\
    \midrule
    Q1& You are asked to write down one sentence at each round.                      & False, \textbf{True} \\
    Q2& You will not know whether you fool the model after you submit your sentence. & \textbf{False}, True \\
    Q3& You may need to wait after you submit your result.                           & False, \textbf{True} \\
    Q4& The goal is to find an example that the model gets 
    right but that another person would get wrong.                                   & \textbf{False}, True \\[1ex]
    Q5& You will be modifying a provided prompt to generate your own sentence.       & False, \textbf{True} \\
    Q6& Here is an example where a powerful model predicts positive sentiment: 
    ``The restaurant is near my place, but the food there is not good at all.''      & \textbf{False}, True \\
    Q7& Suppose the goal is to write a positive sentence that fools the model 
    into predicting another label. The sentence provided is ``That was awful!'', 
    and the model predicts negative. Does this sentence meet the goal of the task?   & Yes, \textbf{No} \\
    Q8& Suppose the goal is to write a negative sentence that fools the model 
    into predicting another label. The sentence provided is ``They did a great 
    job of making me feel unwanted'', and the model predicts positive. Does this 
    sentence meet the goal of the task?                                              & \textbf{Yes}, No \\
    \bottomrule
  \end{tabular}
  \caption{A list of comprehension questions we asked workers to answer correctly with a maximum of 5 retires. Correct answer is bolded.}
  \label{tab:onboarding-quiz}
\end{table*}

\begin{table*}[tp]
  \centering
  \renewcommand{\arraystretch}{1.2}
  \begin{subtable}{1\linewidth}
  \centering
    \begin{tabular}{p{0.6\linewidth} c c}
      \toprule
      &  SST-3 &                     Responses \\
      \midrule
      Moretti 's compelling anatomy of grief and the difficult process of adapting to loss. &       neg &  neu, \textbf{pos, pos, pos, pos} \\
      Nothing is sacred in this gut-buster. &       neg &  neg, neg, \textbf{pos, pos, pos} \\
      \bottomrule
    \end{tabular}
    \caption{All examples for which the SST-3 label is Negative and our majority label is Positive.}
  \end{subtable}

  \vspace{6pt}

  \begin{subtable}{1\linewidth}
  \centering
    \begin{tabular}{p{0.6\linewidth} c c}
      \toprule
      &  SST-3 &                     Responses \\
      \midrule
      ... routine , harmless diversion and little else.  & pos &  mix, mix, \textbf{neg, neg, neg} \\
      Hilariously inept and ridiculous. & pos &  mix, \textbf{neg, neg, neg, neg} \\
      Reign of Fire looks as if it was made without much thought -- and is best watched that way. & pos &  mix, \textbf{neg, neg, neg, neg} \\
      So much facile technique, such cute ideas, so little movie.  & pos &  mix, mix, \textbf{neg, neg, neg} \\
      While there 's something intrinsically funny about Sir Anthony Hopkins saying 'get in the car, bitch,' this Jerry Bruckheimer production has little else to offer  & pos &  mix, \textbf{neg, neg, neg, neg} \\
      \bottomrule
    \end{tabular}
    \caption{All examples for which the SST-3 label is Positive and our majority label is Negative.}
  \end{subtable}

  \vspace{6pt}

  \begin{subtable}{1\linewidth}
    \centering
    \begin{tabular}{p{0.6\linewidth} c c}
      \toprule
      &  SST-3 &                     Responses \\
      \midrule
      Returning aggressively to his formula of dimwitted comedy and even dimmer characters, Sandler, who also executive produces, has made a film that makes previous vehicles look smart and sassy. &  neu &  \textbf{neg, neg, neg, neg, neg} \\
      should be seen at the very least for its spasms of absurdist humor. &  neu &  \textbf{pos, pos, pos, pos, pos} \\
      A workshop mentality prevails. &  neu &  \textbf{neu, neu, neu, neu, neu} \\
      Van Wilder brings a whole new meaning to the phrase ` comedy gag . ' &  neu &  mix, neu, \textbf{pos, pos, pos} \\
      ` They' begins and ends with scenes so terrifying I'm still stunned. &  neu &  neu, neu, \textbf{pos, pos, pos} \\
      Barely gets off the ground. &  neu &  \textbf{neg, neg, neg, neg, neg} \\
      As a tolerable diversion, the film suffices; a Triumph, however, it is not. &  neu &  \textbf{mix, mix, mix, mix}, neg \\
      Christina Ricci comedy about sympathy, hypocrisy and love is a misfire. &  neu &  \textbf{neg, neg, neg, neg, neg }\\
      Jacquot's rendering of Puccini's tale of devotion and double-cross is more than just a filmed opera. &  neu &  neg, neu, \textbf{pos, pos, pos} \\
      Candid Camera on methamphetamines. &  neu &  \textbf{neg, neg, neg}, neu, pos \\
      \bottomrule
    \end{tabular}
    \caption{A random selection of 10 examples for which SST-3 label is Neutral and our validation label is not.}
  \end{subtable}
  \caption{Comparisons between the SST-3 labels and our new validation labels.}
  \label{tab:sst-cmp}
\end{table*}

\begin{table*}[tp]
  \centering
  \renewcommand{\arraystretch}{1.2}
   \begin{subtable}{1\linewidth}
     \centering     
     \begin{tabular}{p{0.6\linewidth} c c}
       \toprule
       Sentence & \ModelOne & Responses \\
       \midrule
       We so wanted to have a new steak house restaurant. &      pos &  \textbf{neu, neu, neu, neu}, pos \\
       As a foodie, I can surely taste the difference. &      pos &  \textbf{neu, neu, neu,} pos, pos \\
       There was however some nice dinner table chairs that I liked a lot for \$35 a piece and for the quality and style this was a very nice price for them. &      pos &  \textbf{pos, pos, pos, pos, pos} \\
       The waitress helped me pick from the traditional menu and I ended up with chilli chicken. &      pos &  neu, neu, \textbf{pos, pos, pos} \\
       I have had lashes in the past and have used some of those Living Social deals. &      pos &  \textbf{neu, neu, neu, neu}, pos \\
       Lots of trash cans. &       neu &  mix, \textbf{neu, neu, neu, neu}\\
       They were out the next day after my call to do the inspection and same for the treatment. &      pos &  mix, \textbf{neu, neu, neu}, pos \\
       When we walked in no one was there to sit us, I waited for a minute and then decided just to take a seat. &      neg &  mix, \textbf{neg, neg, neg, neg} \\
       Driver was amazing! &      pos &  \textbf{pos, pos, pos, pos, pos} \\
       We tried:\textbackslash n\textbackslash nChampagne On Deck - Smooth and easy to drink. &      pos &  neu, neu, \textbf{pos, pos, pos} \\
       \bottomrule
     \end{tabular}
     \caption{Randomly sampled Round~1 train cases.}
   \end{subtable}

   \vspace{6pt}
   
  \begin{subtable}{1\linewidth}
    \centering
     \begin{tabular}{p{0.6\linewidth} c c}
  \toprule
    Sentence & \ModelOne & Responses \\
    \midrule
    The menu was appealing.             & pos &  \textbf{pos, pos, pos, pos, pos} \\
    Our food took forever to get home.  & neg &  \textbf{neg, neg, neg, neg}, neu \\
    Our table was ready after 90 minutes and usually I'd be mad but the scenery was nice so it was an okay time. &  pos &  \textbf{mix, mix, mix, mix}, pos \\
    I left feeling like an initiator in a world of books that I never even knew existed. &      pos &  mix, neu, \textbf{pos, pos, pos} \\
    I decided to go with josh wilcox, only as a last resort &  neu &  mix, neg, \textbf{neu, neu, neu} \\
    : ] Standard ask if the food is a great happening! & neu &  \textbf{neu, neu, neu, neu}, pos \\
    The car was really beautiful. &  pos &  neu, \textbf{pos, pos, pos, pos} \\
    Food is a beast of its own, low-quality ingredients, typically undercooked, and a thin, very simple menu. & neg &  \textbf{neg, neg, neg, neg}, neu \\
    I want to share a horrible experience here. &  neg &  \textbf{neg, neg, neg, neg, neg} \\
    I tried a new place. The entrees were subpar to say the least. & neg &  \textbf{neg, neg, neg, neg, neg} \\
    \bottomrule
  \end{tabular}
     \caption{Randomly sampled Round~2 train cases.}
  \end{subtable}
  \caption{Randomly sampled Round 1 and Round 2 train cases.}
  \label{tab:random-examples}
\end{table*}

\end{document}